\documentclass[10pt, conference, letterpaper]{IEEEtran}

\usepackage{microtype}
\usepackage{graphicx}
\usepackage{booktabs} 

\usepackage{hyperref}
\usepackage{amsmath}
\usepackage{multirow}
\usepackage{xcolor}
\usepackage{subcaption}
\usepackage{tikz}
\usetikzlibrary{matrix}

\usepackage{pifont}
\newcommand{\cmark}{\ding{51}}%
\newcommand{\xmark}{\ding{55}}%

\usepackage{amsmath}
\usepackage{amssymb}
\usepackage{mathtools}
\usepackage{amsthm}

\usepackage[capitalize,noabbrev]{cleveref}

\begin{document}

\title{Towards Fair, Robust and Efficient Client Contribution Evaluation in Federated Learning}
\author{
\IEEEauthorblockN{Meiying Zhang, Huan Zhao, Sheldon Ebron, Kan Yang}\\
\IEEEauthorblockA{Dept. of Computer Science, University of Memphis,  USA.  \\
\{mzhang6, hzhao2, sebron, kan.yang\}@memphis.edu}
}

\maketitle

\begin{abstract}
The performance of clients in Federated Learning (FL) can vary due to various reasons. Assessing the contributions of each client is crucial for client selection and compensation. It is challenging because clients often have non-independent and identically distributed (non-iid) data, leading to potentially noisy or divergent updates. The risk of malicious clients amplifies the challenge especially when there's no access to clients' local data or a benchmark root dataset. In this paper, we introduce a novel method called Fair, Robust, and Efficient Client Assessment (FRECA) for quantifying client contributions in FL. FRECA employs a framework called FedTruth to estimate the global model's ground truth update, balancing contributions from all clients while filtering out impacts from malicious ones. This approach is robust against Byzantine attacks and incorporates a Byzantine-resilient aggregation algorithm. FRECA is also efficient, as it operates solely on local model updates and requires no validation operations or datasets. Our experimental results show that FRECA can accurately and efficiently quantify client contributions in a robust manner.
\end{abstract}


\section{Introduction}

In the context of federated learning (FL), participants or clients actively contribute to the training of a global model by providing local models or updates to the global model trained on their own data. It is important to rigorously quantify the individual contributions of participants, which is an essential step for efficient client selection, fair allocation of profit earned through the FL process, and design of incentive mechanisms aimed at attracting high-valued participants. The assessment of client contributions presents a notable challenge, as data is not directly shared; rather, it is indirectly conveyed through locally trained models utilizing the global model as a foundation. 

Traditional data valuation or pricing methods \cite{chen2019towards, zhan2020learning, sim2020collaborative} are thus not applicable. The degree of contribution is intricately influenced not only by the size and distribution of a client's data but also by factors such as the specific FL task, the initial global model serving as the training basis for the local model, the iteration/round of training in which the client participates, and the collective composition of clients participating in the same training round. Consequently, there is a compelling need for a precise measurement of the contribution made by a client to the global model in each training round, specifically quantifying the impact of each local model or update on the aggregated global model.

Due to various qualities of data and trained local model, it is unfair to treat all the clients equally \cite{nguyen2021flame, bagdasaryan2020backdoor, shen2016auror} or evaluate client contributions based on the size of the training dataset \cite{mcmahan2017communication}. A dishonest client may train the local model over partial datasets or claim a large size of training dataset for more rewards. Some existing client contribution evaluation methods only focus on whether the client has submitted the model updates or whether the norms of model updates are within a threshold \cite{kang2019incentive}. In \cite{wang2019measure}, a deletion-based approach is proposed to evaluate the contribution of an individual client by comparing the accuracy of the global model with and without this client. 

More accurate Shapley value approaches \cite{shapley1953value, song2019profit} model FL as a cooperative game and compute the contribution of each player as the marginal impact on the overall reward which is the accuracy achieved by the global model. However, the Shapley value approach has two significant shortcomings: 
\begin{enumerate}[a)]
    \item It imposes \textit{intensive computational demand}, which stems from the need to reconstruct and evaluate a variety of sub-models. To address this, techniques such as random permutation sampling and group testing have been introduced \cite{wang2020principled, liu2022gtg}. However, these methods only partially mitigate the computational intensity, which becomes particularly burdensome as the number of clients increases; 
    \item The framework requires an \textit{auxiliary validation dataset} to assess the performance of all the sub-models. However, such a validation dataset may not be feasible in many FL applications due to privacy and regulation constraints. 
\end{enumerate}

Another line of approach is distance-based methods \cite{yan2021fedcm, wu2021fast} which assess each client's contribution per FL round based on the distance between the client's local model and the prior global model. 
Unlike Shapley value approaches, distance-based methods do not require extra validation datasets but they face two critical challenges: 
\begin{enumerate}[a)]
    \item \textit{These methods primarily focus on the gap between the recent global model and local updates, \textbf{lacking a solid ground truth} for comparison}. FLTrust \cite{cao2021fltrust} suggests a benign root dataset as a standard, but in federated learning, data privacy issues often hinder its adoption; 

    \item \textit{No defense strategies on the aggregator side are taken into account}. The influence of different Byzantine-resilient strategies during the aggregation process can significantly impact the evaluation of each client's contribution. For instance, in assessing a client's contribution, it may be pertinent to consider factors like the aggregation weights, which could include the ratio of local data samples to the total data samples in algorithms like FedAvg. Consequently, this presents an essential question: should the evaluation of a client's contribution rely solely on the distance between their local model and the aggregated model, or should it be a more nuanced measure that incorporates these aggregation weights?
\end{enumerate}

To address these issues, we introduce a novel method called Fair, Robust, and Efficient Client Assessment (FRECA) for quantifying client contributions in FL. FRECA employs a framework called FedTruth to estimate the global model's ground truth update, balancing contributions from all clients while filtering out impacts from malicious ones. It quantifies each client's contribution based on their share in the gap distance between the estimated and actual ground truth. This approach is robust against Byzantine attacks and incorporates a Byzantine-resilient aggregation algorithm. FRECA is also efficient, as it operates solely on local model updates and requires no validation operations or datasets. Our experimental results show that FRECA can accurately and efficiently quantify client contributions in a robust manner.




The contributions of this paper are summarized as follows: 
\begin{itemize}
    \item We propose a novel method called Fair, Robust, and Efficient Client Assessment (FRECA) for quantifying client contributions in FL. FRECA employs a framework called FedTruth to estimate the global model's ground truth update, balancing contributions from all clients while mitigating impacts from malicious ones.       
    \item To the best of our knowledge, this is the first contribution evaluation method to incorporate defense mechanism against malicious clients. This approach is robust against Byzantine attacks and also efficient, as it operates solely on local model updates and requires no validation operations or datasets.
    \item Our experimental results show that FRECA can accurately and efficiently quantify client contributions in a robust manner.
\end{itemize}

The remainder of this paper is organized as follows: In Section \ref{sec:relatedwork}, we present the related work of client contribution evaluation and Byzantine-resilient aggregation algorithms in FL. 
Section \ref{sec:preliminaries} describes the problem formulation of federated learning, existing client contribution assessment methods, and FedTruth framework to estimate ground truth of the global model update. In Section \ref{sec:freca}, we present our efficient and robust client assessment method FRERA. Section \ref{sec:evaluation} provides experimental evaluation. Finally, Section \ref{sec:conclusion} concludes the paper.

\begin{table*}[!h]
    \centering
    \caption{Comparison of Client Assessment Goals}
    \small
    \begin{tabular}{c|c|c|c|c}
        \hline\hline 
        \textbf{Scheme} & \textbf{No Validation Dataset} & \textbf{Efficient} & \textbf{Byzantine-resilient} & \textbf{Attack Detection} \\ \hline
        Shapley Value approach \cite{song2019profit}  & \xmark & \xmark & \xmark & \xmark \\ \hline
        LOO approach \cite{wang2019measure} & \xmark  & \cmark & \xmark & \xmark \\ \hline
        Distance-based approach \cite{yan2021fedcm} & \cmark  & \cmark & \xmark & \xmark \\ \hline
        Our FRECA &  \cmark & \cmark  & \cmark  & \cmark \\ \hline \hline
    \end{tabular}
    \label{table:comparison}
\end{table*}


\section{Related Work}\label{sec:relatedwork}
Many methods have been proposed to measure the contribution of a client in FL, which usually fall in two directions:
\textit{Shapley Value Approaches} and \textit{Distance-based Approaches}. 

\subsection{Shapley Value Apporach}
The concept of the Shapley value, eloquently described by Shapley in his pioneering work \cite{shapley1953value}, serves as an equitable framework for gauging contribution. It calculates the marginal contribution, delineating the variance in overall rewards when a participant either engages in or refrains from a particular activity. In a series of insightful papers, Jia \textit{et al.} delineated methodologies to compute the Data Shapley more efficiently. Their notable strategies encompass utilizing Locality Sensitive Hashing (LSH) for approximations in KNN settings \cite{jia2019efficient} and capitalizing on the inherent sparsity of Shapley values to reduce the demands of model evaluations \cite{jia2019towards}.

Inspired by this foundational concept, subsequent researchers have attempted to envision each client in the federated learning landscape as an individual `player.' The aim here is to assess the influence of each client on the overall model's performance. Specifically, the focus is on evaluating the model's performance shift when a particular client becomes part of or exits from a group. This aggregate assessment is carried over myriad potential coalitions that exclude the client under consideration. A vivid illustration of this approach is presented in \cite{song2019profit}, where the Contribution Index (CI) is introduced, echoing the principles of the Shapley value. In a parallel vein, \cite{wang2020principled} introduces the Federated Shapley value (Federated SV) that uniquely considers the chronological order of client participation, shedding light on the intricate nuances of data value. Both CI and Federated SV not only adhere to the fairness principles of the Shapley value but also offer feasible computational methods through approximation algorithms.

\subsection{Distance-based Approaches} 
In \cite{yan2021fedcm}, researchers introduced a novel methodology where the contribution of each client is determined every round for every layer of the model parameters. This evaluation utilizes the `attention weight' (effectively the aggregation weight) corresponding to the local model update. The attention weight is discerned based on the divergence between a client's local model and the global model from the previous round. A foundational presumption of this strategy is the belief that the larger the influence a client exerts on the global model, the more significant their contribution. Yet, real-world complexities might challenge this presumption. Similarly, \cite{wu2021fast} measures client contribution by inspecting the angular difference between local and global loss function gradients, postulating that a smaller angle signifies a more pronounced contribution to the global model update.

Distance-based methodologies present a distinct advantage over Shapley value techniques, in that they \textit{{eliminate the need to assess model performance using supplementary validation datasets}}. However, they do grapple with a notable challenge: their primary emphasis is on determining the distance between the latest global model and the local model updates. This issue stems from {\textit{the lack of a definitive ``ground truth" for the global model}}, which would otherwise serve as a standard for distance measurements. One proposed solution, as highlighted in FLTrust \cite{cao2021fltrust}, involves using a benign root dataset as this standard. Yet, in the realm of federated learning, this solution often proves untenable due to prevailing data privacy and regulatory hurdles.

\subsection{Byzantine-Resilient Aggregation Algorithms}
Byzantine attack is a common attack in federated learning that aims to make the global model converged to a sub-optimal model by arbitrarily altering local model updates. Types of Byzantine attack include model-boosting attack \cite{bhagoji2019analyzing}, Gaussian noise attack \cite{blanchard2017machine}, and constraint-and-scaling attack  \cite{bagdasaryan2020backdoor}. Several aggregation methods are proposed \cite{ blanchard2017machine, yin2018byzantine, cao2021fltrust} to defend against this attack. \emph{Krum}~\cite{blanchard2017machine} selects the local model from one 'best' client as the global model for each round, thus ignoring contributions from other clients. \emph{Trimmed Mean}~\cite{yin2018byzantine} tries to remove malicious clients by trimming outliers from local models, but in this way, benign models trained on underrepresented data may also be removed. In \textit{FLTrust} \cite{cao2021fltrust}, aggregation weights are estimated based on the similarity between each model update with a \emph{ground-truth model update} which is trained by the aggregator using a benign root dataset. However, this benign root dataset may not be practical in many applications.

Table \ref{table:comparison} lists the comparison between our proposal FRECA and the existing approaches. 

\section{Preliminaries}\label{sec:preliminaries}

\subsection{Federated Learning}
A general FL system consists of an aggregator and a set of clients $S$. Let $\mathcal{D}_k$ be the local dataset held by the client $k~(k\in S)$. The typical FL goal \cite{mcmahan2017communication} is to learn a model collaboratively without sharing local datasets by solving 
\begin{equation}
\begin{split}
    \min_{w} F(w)  &= \sum_{k\in S} p_k\cdot F_k(w), \\
    ~s.t.~\sum_{k\in S} p_k  & = 1 ~(p_k\geq 0),
\end{split}
\end{equation}
where 
\[
F_k(w) = \frac{1}{n_k}\sum_{j_k = 1}^{n_k} f_{j_k}(w; x^{(j_k)}, y^{(j_k)})
\]
is the local objective function for a client $k$ with $n_k = |\mathcal{D}_k|$ available samples. $p_k$ is usually set as $p_k = n_k/\sum_{k\in S} n_k$ (e.g., FedAvg \cite{mcmahan2017communication}). The FL training process usually contains multiple rounds, and a typical FL round consists of the following steps: 
\begin{enumerate}
    \item \textit{client selection and model update}: a subset of clients $S_t$ is selected, each of which retrieves the current global model $w_t$ from the aggregator.
    \item \textit{local training}: each client $k$ trains an updated model $w^{(k)}_t$ with the local dataset $\mathcal{D}_k$ and shares the model update $\Delta_t^{(k)} = w_t - w_t^{(k)}$ to the aggregator.
    \item \textit{model aggregation}: the aggregator computes the global model updates as $\Delta_t = \sum_{k\in S_t} p_k \Delta_t^{(k)}$ and update the global model as $w_{t+1} = w_t - \eta \Delta_t$, where $\eta$ is the server learning rate.
\end{enumerate}

{FedAvg ~\cite{mcmahan2017communication}} is the original aggregation rule, which generates a representative global model after receiving the local models from trustworthy (i.e., benign) participants. This algorithm averages all local model weights selected based on the number of samples the participants used. FedAvg has been shown to work well when all the participants are benign clients, but is vulnerable to model poisoning attacks such as Byzantine attack.

\subsection{Client Assessment}

Shapley value is the most commonly used state-of-the-art method for client contribution assessment in federated learning. In game theory, a player's Shapley value is a weighted sum of marginal contributions of all possible coalitions (group of players), where marginal contribution is the difference in total rewards between the player joining and not joining the coalition \cite{shapley1953value, ghorbani2019data}. In a FL setting, Shapley value-based contribution \cite{song2019profit} is defined as follows. 

\begin{equation}\label{sv}
    SV_t(k) = C\sum_{S \subseteq S_t \setminus \{k\}} \frac{U(M_{S \cup \{k\}})-U(M_S)}{\binom{|S_t|-1}{S}}
\end{equation}
where $t$ denotes an FL aggregation round and $k$ denotes a client, $C$ is a constant, and $U(M_S)$ is a utility function of a model $M$ trained on a group of clients $S$. The utility function can be the accuracy of the model evaluated on a validation dataset. 

Another way to measure the contribution of a client is the deletion method or leave-one-out (LOO) method \cite{wang2019measure}, which calculates the change in model performance when the client is removed from the client group participating in the same round. Using the same notation as in Equation \ref{sv}, LOO contribution can be expressed as follows. 

\begin{equation}\label{loo}
    LOO_t(k) = U(M_{S_t}) - U(M_{S_t \setminus \{k\}})
\end{equation}

In this paper, we use the above two contribution assessment methods as baselines to compare our proposed method. 

\subsection{FedTruth: Estimating the ground-truth of global model update}\label{sec:FedTruth}
Without a benign root dataset, it is challenging to obtain the \emph{ground-truth model update} among all the local model updates in an FL round. In \cite{ebron2023fedtruth}, the authors propose a new model aggregation algorithm, namely FedTruth, which enables the aggregator to uncover the truth among all the received local model updates. The basic idea of FedTruth is inspired by truth discovery mechanisms \cite{li2016conflicts, yin2008truth, li2015discovery, ouyang2016aggregating}, which are developed to extract the truth among multiple conflicting pieces of data from different sources under the assumption that the source reliability is unknown a priori. The \emph{ground-truth model update} is computed as the weighted average of all the local model updates with dynamic aggregation weights for each round. The weights in FedTruth are dynamically chosen based on the distances between the estimated truth and local model updates, following the principle that higher weights will be assigned to more reliable clients. 

Suppose the aggregator receives $n_t$ different model updates $\Delta_{t}^{(1)}, \cdots$, $\Delta_{t}^{(n_t)}$ in FL round $t$. To find the global update $\Delta^*_{t}$, we 
formulate an optimization problem aiming at minimizing the total distance between all the model updates and the estimated global update: 
\begin{equation}\label{equ:fedtruth}
\begin{split}
    \min_{\Delta^*_{t}, {\bf p_{t}}} D(\Delta^*_{t}, {\bf p_{t}}) & = \sum_{k=1}^{n_t} g(p_{t}^{(k)})\cdot d(\Delta^*_{t}, \Delta_{t}^{(k)})\\
    ~~~ s.t.~~~\sum_{k=1}^{n_t} p_{t}^{(k)}  & = 1
\end{split}
\end{equation}
where $d(\cdot)$ is the distance function and $g(\cdot)$ is a non-negative coefficient function. $p_{t}^{(k)}$ is the performance of the local model $\Delta_{t}^{(k)}$ which is calculated based on the distance. 


To solve this optimization problem, FedTruth iteratively computes the estimated truth ${\Delta}^*_t$ and the performance values ${\bf p^t}$ using coordinate descent approach \cite{bertsekas1997nonlinear}. 

\textit{Updating Aggregation Weights:}
Once the truth $\Delta^*_{t}$ is fixed, FedTruth first calculates the performance of each model update $\{p_t^{(k)}\} (k=1,\cdots, n_t)$ as 
\begin{equation}\label{equ:p}
    p_t^{(k)} = {d(\Delta^*_{t}, \Delta_{t}^{(k)})}/{\sum\limits_{k'=1}^{n_t} d(\Delta^*_{t}, \Delta_{t}^{(k')})}.
\end{equation}
Then, the aggregation weights can be updated as 
\begin{equation}\label{equ:a}
    a_t^{(k)} = \frac{g(p_t^{(k)})}{\sum_{k=1}^{n_t} g(p_t^{(k)})}.
\end{equation}

\textit{Updating the Truth:}
Based on the new aggregation weights $\{a_t^{(1)}, \cdots, a_t^{(n_t)}\}$, the truth of global update can be estimated as 
\begin{equation}
    \Delta^*_{t} = \sum_{k=1}^{n_t} a_t^{(k)}\cdot  \Delta_{t}^{(k)}
\end{equation}

\begin{figure*}[!t]
    \centering
    \includegraphics[width=.85\linewidth]{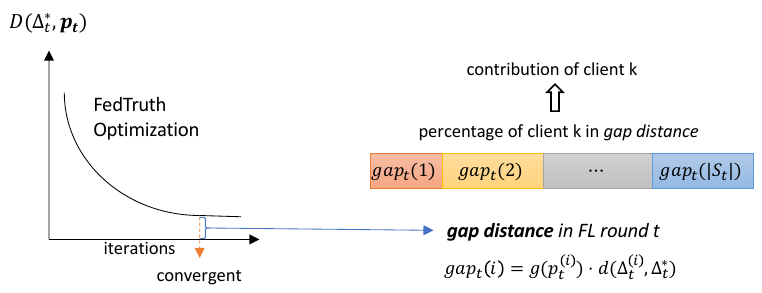}
    \caption{Gap Distance between all model updates and the estimated global model update}
    \label{fig:gap}
\end{figure*}

The global model update and aggregation weights will be updated iteratively until convergence criteria are met. It is easy to see that the longer the distance between the local model update and the estimated truth, the smaller aggregation weight will be assigned in calculating the truth. This principle can eliminate the impacts of malicious model updates and keep certain contributions from a benign outlier model update.

\section{FRECA: A Fair, Robust and Efficient Client Assessment Method}\label{sec:freca}

To ensure fair client assessment in FL, we introduce two distinct metrics: the Client Performance Evaluation Metric and the Net Contribution Evaluation Metric. The Client Performance Evaluation Metric assesses the divergence between an individual client's model outputs and the ground truth. Conversely, the Net Contribution Evaluation Metric quantifies the extent of an individual client's contributions that are incorporated into the overall global model.

\subsection{Client Performance Evaluation Metric}
In the FedTruth aggregation algorithm, larger aggregation weights will be assigned to the model updates closer to the global model update, so the aggregation weight can somehow reflect the reliability of the clients. We will use this aggregation weight (AW) to evaluate the client performance. 

According to Equ.\ref{equ:p}, the performance of each model is calculated based on the distance between the local model and the estimated truth of the global model. For example, the distance function $d(\cdot)$ can be expressed as: 
\begin{itemize}
    \item Euclidean distance:
\begin{equation}\label{equ:distance1}
    d_l(\Delta^*_{t}, \Delta_{t}^{(k)}) = ||\Delta^*_{t} - \Delta_{t}^{(k)}||
\end{equation}

    \item Angular distance:
\begin{equation}\label{equ:distance2}
    d_a(\Delta^*_{t}, \Delta_{t}^{(k)}) = arccos(S_c(\Delta^*_{t} - \Delta_{t}^{(k)}))/\pi
\end{equation}
where $S_c$ is the cosine similarity. 

    \item Hybrid distance: 
\begin{equation*}\label{equ:distance3}
    d(\Delta^*_{t}, \Delta_{t}^{(k)}) = \alpha \cdot d_l(\Delta^*_{t}, \Delta_{t}^{(k)}) + (1 - \alpha) \cdot d_a(\Delta^*_{t}, \Delta_{t}^{(k)})
\end{equation*}    
where $\alpha \in [0, 1]$ is a combination weight. 
\end{itemize}

The aggregation weight is calculated based on the regulation function $g(\cdot)$ as shown in Equ. \ref{equ:a}. In order to guarantee the convergence of FedTruth, the authors have shown that this regulation function should be monotonous and differentiable in the aggregation weight domain.  Moreover, according to the principle of truth discovery, $g(\cdot)$ should be a decreasing function. Some simple but effective coefficient functions are as follows: 
\begin{equation}
     g(p_t^{(k)}) = 1/p_t^{(k)}~~\text{or}~~ g(p_t^{(k)}) = - \log(p_t^{(k)}).
\end{equation}

Therefore, \textit{the client performance can be quantified as the aggregation weight of the converged iteration of FedTruth}. 
In our experiments, we choose the Euclidean distance function and the $g(p_t^{(k)}) = 1/p_t^{(k)}$ as the regulation function.


\subsection{Net Contribution Evaluation Metric}
If the aggregation is the simple average of all the local models. The net contribution is the same as the client performance (i.e., aggregation weights). However, in FedTruth, the aggregation weights are dynamically calculated during the estimation of the truth of the global model in each FL round. To answer the question that ``should the evaluation of a client's contribution rely solely on the distance between their local model and the aggregated model, or should it be a more nuanced measure that incorporates these aggregation weights?''. We propose a novel net contribution evaluation metric that counts both the aggregation weights and the model distance contributing to client contribution.

As shown in Fig. \ref{fig:gap}, to evaluate the net contribution of each client in each FL round $t$, we will first define a \textit{\textbf{gap distance}} of this round, which measures the gap distance between all the model updates and the converged ground truth of the global model update: 
\begin{equation}
    gap_t(S_t) = \sum_{i\in S_t} gap_t(i) = \sum_{i\in S_t} g(p_t^{(i)})\cdot d(\Delta_t^{(i)}, \Delta^*_t)
\end{equation}
where $S_t$ is the set of participating clients in round $t$, and $\Delta^*_t$ is the converged global model update.

We further compute how much percentage of a client $k$ contributes to this gap distance by considering both the aggregation weights and the distance between its local model updates and the estimated global model update: 
\begin{equation}\label{equ:loss}
	\ell_t^{(k)} = \frac{g(p_t^{(k)})\cdot d(\Delta_t^{(k)}, \Delta^*_t)}{\sum_{i\in S_t} g(p_t^{(i)})\cdot d(\Delta_t^{(i)}, \Delta^*_t)},
\end{equation}

With no access to any individual/global dataset that can be used to evaluate the accuracy of the global model, we define \textbf{net contributions based on the percentage in the gap distance}, i.e., 
\textit{a client contributes more to the global model if it has less percentage in the gap distance.}

Specifically, given a set of percentages $\{\ell_t^{(k)}\}_{k\in S_t}$ where $\sum_{k\in S_t} \ell_t^{(k)} = 1$, client contributions $\{\mathcal{C}_t^{(k)}\}_{k\in S_t}$ can be calculated by solving the following linear equation:
\begin{equation}\label{netcontrib}
\sum_{i\in S_t} \mathcal{C}_t^{(k)} = 1~~~~~\text{and}~~~~\frac{\ell_t^{(i)}}{\ell_t^{(k)}} = \frac{\mathcal{C}_t^{(k)}}{\mathcal{C}_t^{(i)}}, \forall i, k\in S_t
\end{equation}

For example, suppose the percentages in the gap distance of four clients are
\[
\ell_t^{(1)} = 0.1, \ell_t^{(2)} = 0.2, \ell_t^{(3)} = 0.3, \ell_t^{(4)} = 0.4,
\]
client net contributions can be calculated as 
\[
\mathcal{C}_t^{(1)} = 12/25, \mathcal{C}_t^{(2)} = 6/25, \mathcal{C}_t^{(3)} = 4/25, \mathcal{C}_t^{(4)} = 3/25.
\]

\section{Experimental Results}\label{sec:evaluation}
\subsection{Settings}
We implement FL with FedTruth aggregation algorithm on 8 clients using MNIST, CIFAR-10 and FashionMNIST datasets. For each, we trained a CNN model for 10-30 aggregation rounds, with 10 local epochs, a batch size of 64 and a learning rate of 0.001. The machine learning models were implemented using PyTorch framework, and the experiments were run on the Google Colab platform using GPU backend resources with 51.0GB System RAM, 15.0GB GPU RAM, and 166.8GB Disk. 

For each client in each round, we computed Net Contribution (FRECA Net) (Equation \ref{netcontrib}) and Aggregation Weight (AW) (Equation \ref{equ:a}) as client performance metric (FRECA AW). As baselines, we computed Shapley Value (SV) (Equation \ref{sv}) and Leave-One-Out (LOO) (Equation \ref{loo}), scaling the values to the range of 0 to 1 using min-max scaling and Softmax function. We averaged these metrics across rounds to obtain final contribution metrics for each client. 

We observed that contribution metrics tend to stabilize within the first 10 aggregation rounds, thus FL was performed for 10-30 rounds without considering the convergence of the global model.  

We present our client assessment results for 5 cases regarding client data distribution and Byzantine attack scenarios: 
\begin{itemize}[noitemsep,nolistsep]
    \item Case 1: non-iid setting, each client having a different number of labels in their data
    \item Case 2: non-iid setting, each client having 1 or 2 labels in their data  
    \item Case 3: iid setting, 1 attacker among clients
    \item Case 4: iid setting, 2 attackers among clients
    \item Case 5: iid setting, 3 attackers among clients (in appendix)
\end{itemize}

Non-iid setting means each client does not have data samples for all labels, iid means each client have  samples for all labels and the samples are distributed uniformly across all labels. 

\begin{figure*}[!t]
    \centering
    \includegraphics[width=1\textwidth]{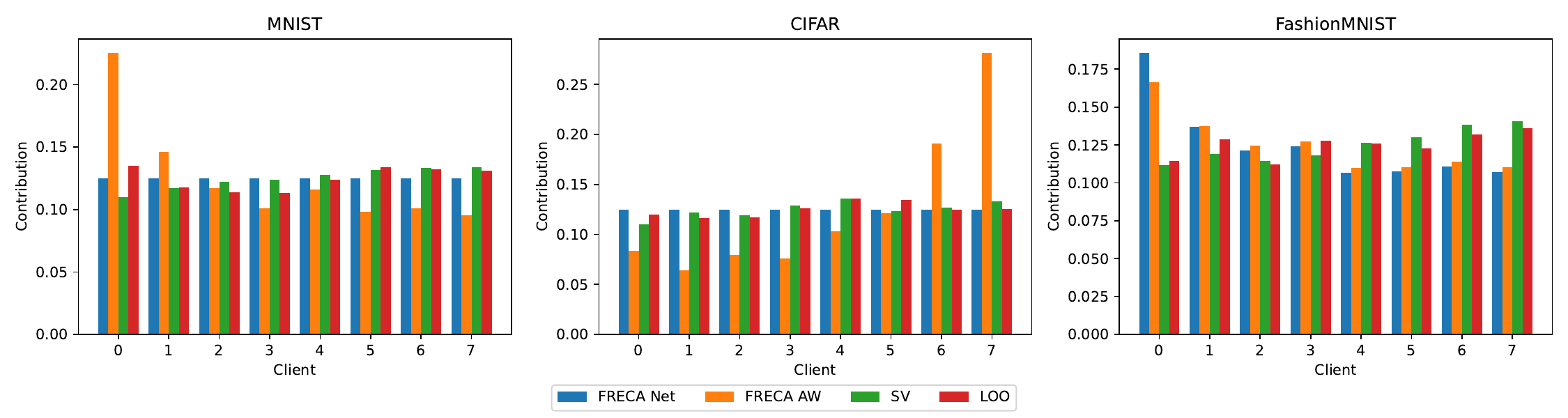}
    \caption{Contribution for Case 1: \# labels for Client 0-7: 1, 2, 3, 4, 6, 8, 9, 10}
    \label{fig:case1}
\end{figure*}

\begin{figure*}[]
    \centering
    \includegraphics[width=1\textwidth]{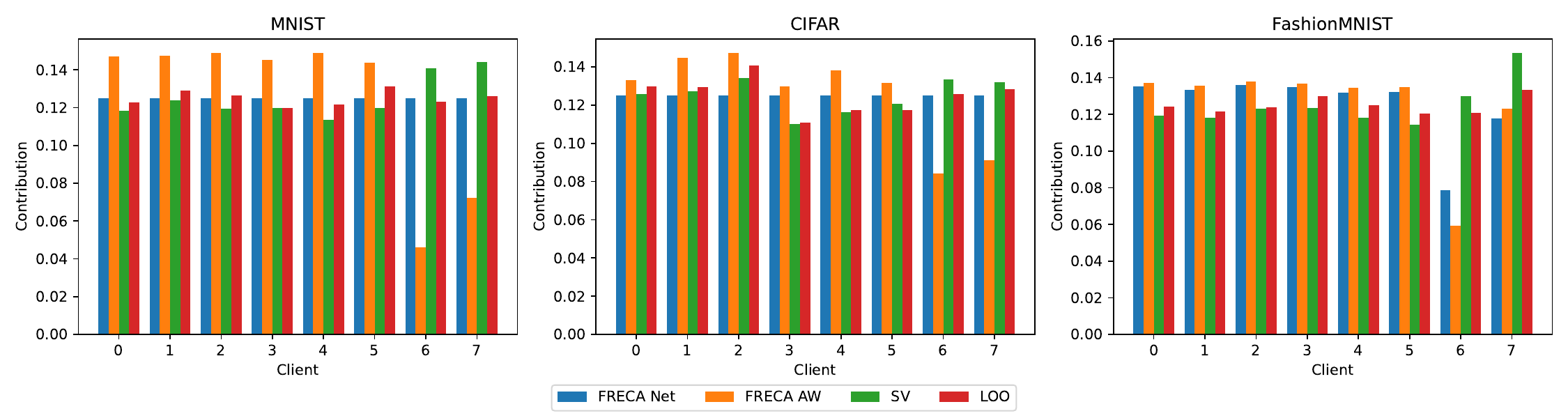}
    \caption{Contribution for Case 2: \# labels for Client 0-7: 1, 1, 1, 1, 1, 1, 2, 2}
    \label{fig:case2}
\end{figure*}

\begin{figure*}[]
    \centering
    \includegraphics[width=1\textwidth]{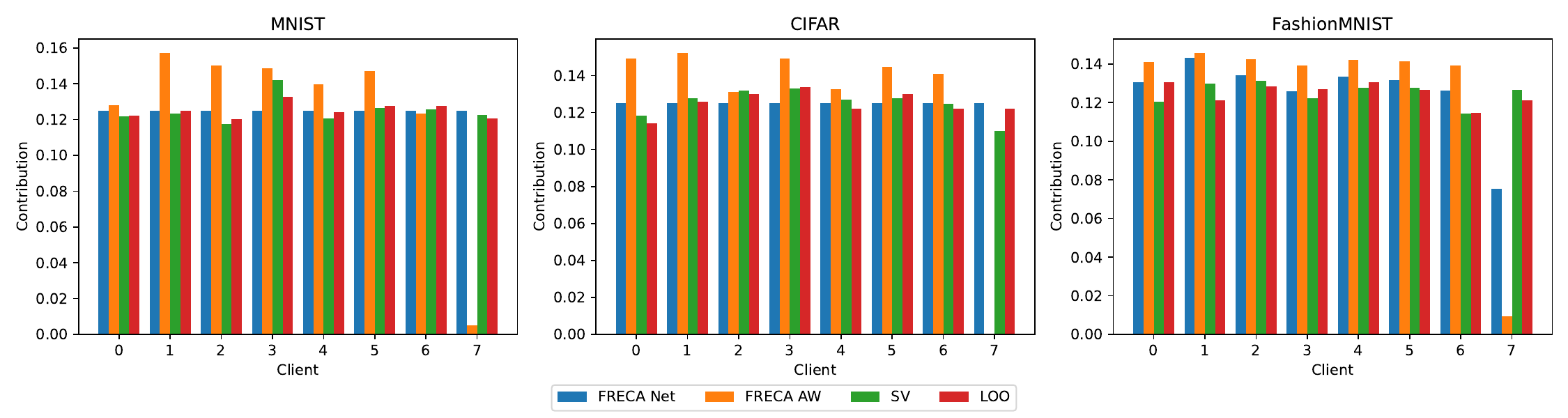}
    \caption{Contribution for Case 3: iid, 1 attacker}
    \label{fig:case3}
\end{figure*}

\begin{figure*}[]
    \centering
    \includegraphics[width=1\textwidth]{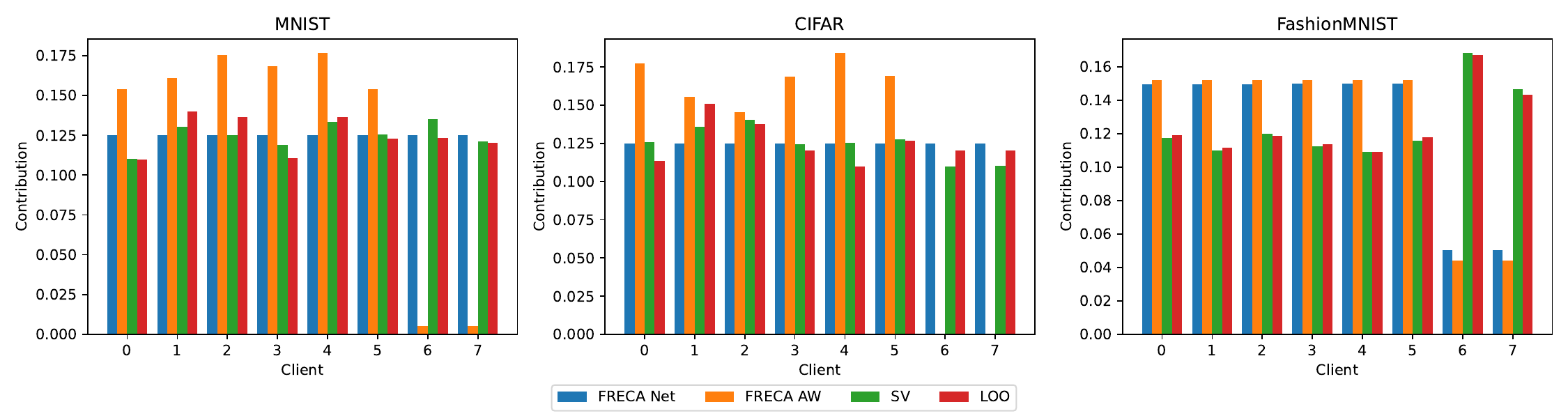}
    \caption{Contribution for Case 4: iid, 2 attackers}
    \label{fig:case4}
\end{figure*}

\subsection{Case 1: non-iid setting, each client having different number of labels in their data}
In Case 1, the number of labels is 1, 2, 3, 4, 6, 8, 9, 10 for Clients 0 to 7, i.e., Client 0 has data samples with 1 label, Client 1 has 2 labels, and so on. To total data size is the same for each client. 

Fig. \ref{fig:case1} shows 4 contribution metrics for each client with 3 datasets. Notice that FRECA Net (blue) is similar to SV (green) or LOO (red) in most cases, indicating our method computes the same assessment as SV/LOO in much less time (see Fig. \ref{fig:time}). FRECA AW (orange) mostly aligns with the net contribution with a few exceptions which can be partially explained by the composition of different labeled data within all 8 clients. For example, Client 0 in MNIST case provides data samples with one label that takes up about 70\% of total samples for this label, which may have led to a higher aggregation weight. Clients 6 and 7 in CIFAR may have contributed more to the global model by providing data more complete in labels. 

\subsection{Case 2: non-iid setting, each client having 1 or 2 labels in their data}
In this case, 6 clients have data with 1 label, 2 clients have data with 2 labels, the sample size per label being the same across clients. Labels are assigned to clients in a non-replacement manner such that all 10 labels are covered. In Fig. \ref{fig:case2}, we see roughly similar values for net contribution, SV and LOO indicating similar contributions from different clients. The AW values are relatively low for the last 2 clients, recognizing the outliers among the clients. This outlier-identifying ability of AW can be utilized to detect malicious clients as detailed in Case 3. 

\subsection{Case 3: iid setting, 1 attacker among clients}
In this setting, each client has the same sample size, distributed uniformly across all labels. One client (Client 7) commits a attack to the global model by amplifying its local model parameters by a factor (e.g., 10). It is obvious from Fig. \ref{fig:case3} that this malicious client is successfully identified by FRECA AW assigning near-zero values to this client. This significantly small AW diluted the impact of the malicious amplified model, resulting in a net contribution similar to other clients. The SV and LOO, on the other hand, were not able to detect the malicious client. 

\begin{figure}[!h]
    \centering
    \includegraphics[width=0.5\textwidth]{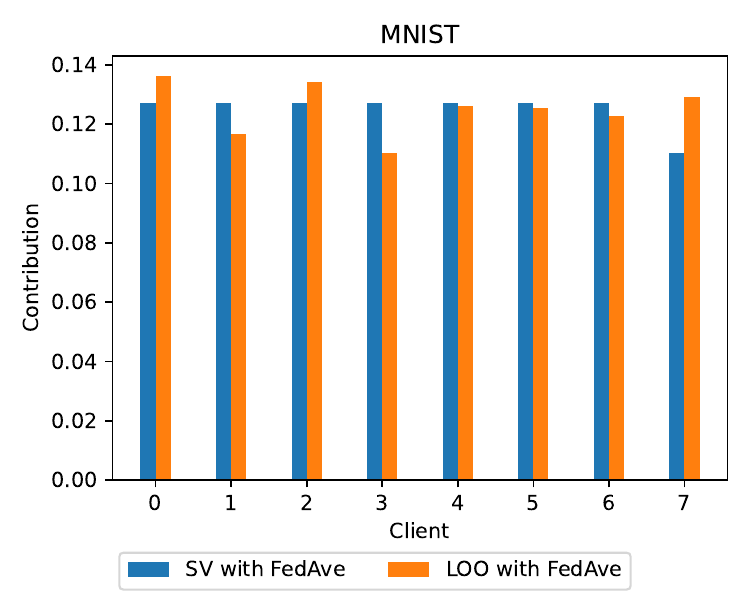}
    \caption{SV \& LOO with FedAve for Case 3}
    \label{fig:case3_fedave}
\end{figure}

\textbf{SV \& LOO with FedAve}:
For comparison, we conducted another FL experiment using FedAve aggregation algorithm, instead of FedTruth, with MNIST dataset and the same set of clients, and computed SV and LOO values as shown in Fig. \ref{fig:case3_fedave}. It can be seen that LOO has no power to detect the attacker, SV did assign a slightly smaller value but it is hardly a significant difference. Thus, in an attack scenario with a traditional aggregation without any defense mechanism, SV or LOO cannot detect the attack. 

\subsection{Case 4: iid setting, 2 attackers among clients}
In Case 4, client data settings are the same as in Case 3, but this time, there are two attackers: Client 6 and 7. Similar to Case 3, we see a stark difference in AW between attackers and normal clients, successfully identifying the attacks with MNIST and CIFAR. With FashionMNIST dataset, SV and LOO assign higher values to the attackers which can be disastrous, while on the contrary, FRECA Net and AW both give higher values to non-attackers and much lower values to attackers. With FRECA Net and AW combined, we can be sure that the last two clients have significantly lower contribution than others. 


\subsection{Case 5: iid setting, 3 attackers among clients}
This is Case 4 with one more attacker (Client 5). Similar patterns are observed to those of 2-attacker case. In MNIST and CIFAR cases, attackers' AW values are close to 0, revealing their maliciousness. In the FashionMNIST case, attackers have higher values of SV and LOO but lower values of both FRECA Net and AW. Although the difference between malicious and normal clients is not as distinct as in MNIST or CIFAR case, we can still identify the last 3 clients as low-contribution participants based on FRECA Net and AW values. 

\begin{figure*}[!h]
    \centering
    \includegraphics[width=1\textwidth]{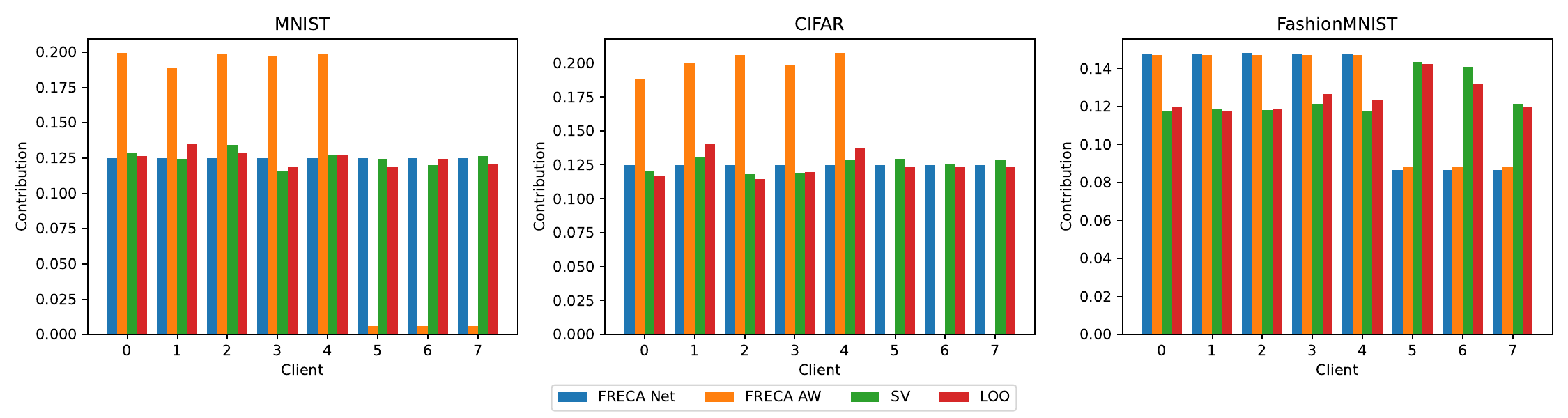}
    \caption{Contribution for Case 5: iid, 3 attackers}
    \label{fig:case5}
    \vspace{-2pt}
\end{figure*}

\subsection{Time Efficiency}
\begin{figure}[!t]
    \centering
    \includegraphics[width=0.5\textwidth]{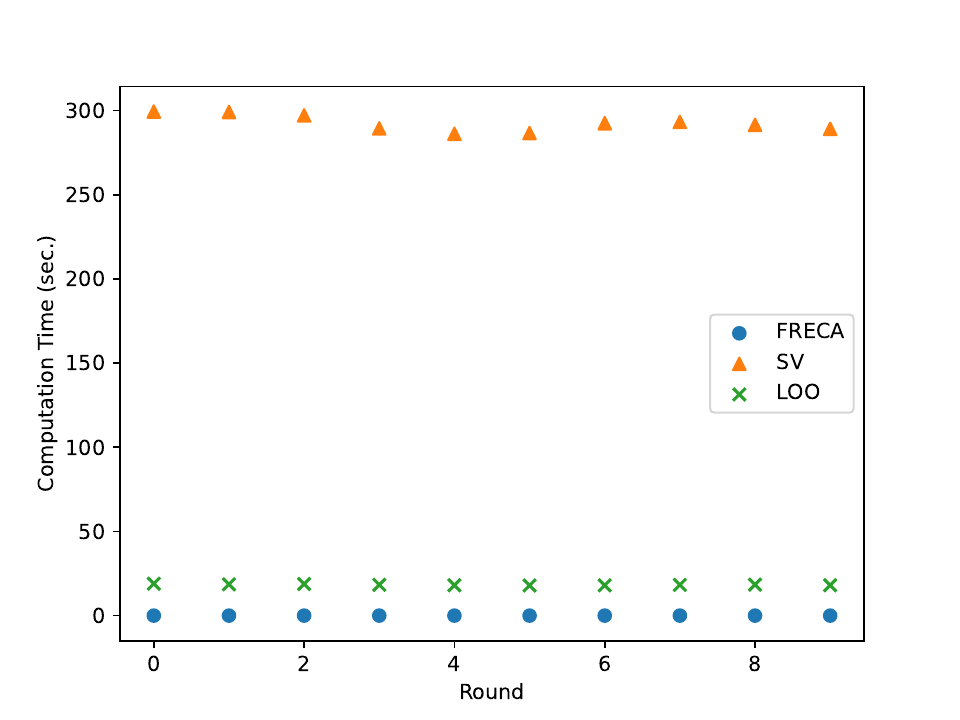}
    \caption{Computation Time Comparison}
    \label{fig:time}
\end{figure}

The time taken to compute FRECA Net and FRECA AW (combined as FRECA for computing both metrics), SV, and LOO is depicted in Fig. \ref{fig:time}, for 10 rounds of FL with 8 clients. The definitions of these metrics tell us the time complexity is $O(2^n)$ for SV and $O(n)$ for LOO and FRECA, with $n$ being the number of clients. As is expected, the computation time for SV is the highest, averaging to 292 sec. per round. This is due to the exhaustive evaluation of aggregated models on the validation dataset for all possible combinations of clients. The average time for LOO and FRECA is 18 sec. and 1.5 sec., respectively. With the same theoretical time complexity $O(n)$, LOO is still much slower than our method because of the evaluation on the validation dataset. 

\section{Conclusion}\label{sec:conclusion}
Quantifying the contribution made by a client to the global model in each training round in FL is crucial to client selection and compensation. Existing client contribution evaluation methods incur high computation cost, require a global dataset to evaluate the accuracy of the global model, or lack a solid ground truth for comparison. Also, they do not consider defense strategies on the aggregator side taken to defend against malicious clients. To address these issues, we introduce a novel method FRECA to quantify client contributions in FL, employing FedTruth framework to estimate the global model's ground truth update, which can balance contributions from all clients while filtering out impacts from malicious ones. This approach is robust against Byzantine attacks as it incorporates a Byzantine-resilient aggregation algorithm, and efficient as it operates solely on local model updates and requires no validation datasets. We show through our experimental results that FRECA can accurately and efficiently quantify client contributions in a robust manner.



\bibliographystyle{IEEEtranS}
\bibliography{Reference.bib}




\end{document}